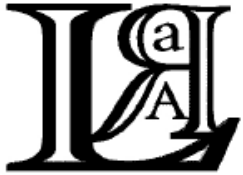

The Artificial Life and Adaptive Robotics Laboratory
ALAR Technical Report Series

# Robustness and Adaptability Analysis of Future Military Air Transportation Fleets

Slawomir Wesolkowski, Michael Mazurek, James Whitacre, Hussein Abbass, and Axel Bender

TR-ALAR-200904006

# Robustness and Adaptability Analysis of Future Military Air Transportation Fleets

**Slawomir Wesolkowski[1], Michael Mazurek[1], James Whitacre[2], Hussein Abbass[2], and Axel Bender[3]**

[1]Centre for Operational Research and Analysis, Defence Research and Development Canada, Ottawa, Canada

[2]School of Information Technology and Electrical Engineering; University of New South Wales at the Australian Defence Force Academy, Canberra, Australia

[3]Land Operations Division, Defence Science and Technology Organisation; Edinburgh, Australia

Email: {s.wesolkowski@ieee.org, J.Whitacre@adfa.edu.au, H.Abbass@adfa.edu.au, Axel.Bender@dsto.defence.gov.au}

## Abstract

Making decisions about the structure of a future military fleet is challenging. Several issues need to be considered, including multiple competing objectives and the complexity of the operating environment. A particular challenge is posed by the various types of uncertainty that the future holds. It is uncertain what future events might be encountered and how fleet design decisions will influence these events. In order to assist strategic decision-making, an analysis of future fleet options needs to account for conditions in which these different uncertainties are exposed. It is important to understand what assumptions a particular fleet is robust to, what the fleet can readily adapt to, and what conditions present risks to the fleet. We call this the analysis of a fleet's strategic positioning. Our main aim is to introduce a framework that captures information useful to a decision maker and defines the concepts of robustness and adaptability in the context of future fleet design. We demonstrate our conceptual framework by simulating an air transportation fleet problem. We account for uncertainty by employing an explorative scenario-based approach. Each scenario represents a sampling of different future conditions and different model assumptions. Proposed changes to a fleet are then analysed based on their influence on the fleet's robustness, adaptability, and risk to different scenarios.

## 1. INTRODUCTION

Some of the most daunting societal challenges are in large part due to our uncertainty about the future. For instance, the difficulty of planning and decision making under uncertainty is evident in domains such as financial market regulation and climate change. The challenges faced by defence planners are equally daunting. As articulated in [1], "*systems design in a context where the demands and tasks, threats and opportunities are both unpredictable and in rapid flux is a serious challenge. Strategic planning is often conducted in the face of massive uncertainty across many dimensions* [2]. *The increasing rate of technological change provides new opportunities equally to the defence force and to potential adversaries.*"

In order to deal with these challenges we need to understand where and how uncertainty becomes important to our choices. We need a methodology that captures relevant types of uncertainty and accounts for uncertainty during our evaluation of options. Effective options (solutions) should be robust under varying conditions and able to adapt to unanticipated threats and opportunities. However, due to multiple competing objectives, as well as the distinctiveness of the plausible environments we might face, no single optimal solution exists. As a result, it is important that we have methods at hand that help us understand better the conditions under which a particular option is robust, when it can be adapted to adequately address emerging challenges, and under which an option poses a significant risk. Similar concepts have been advocated in [3,4].

The next section discusses some of the underlying causes of uncertainty and how uncertainty might be captured using scenario-based or ensemble-based computational techniques. This is followed by a description of the air transportation fleet problem, and the scenario generation process for this problem. In Section 4, a multi-objective optimization algorithm used to search for candidate solutions is described. In Section 5 we outline how options can be analyzed to better understand their particular strengths and weaknesses. Resulting options are then analysed in Section 6 and conclusions are drawn in Section 7.

## 2. MODELLING UNCERTAINTY

Uncertainty can be modelled using various approaches, and each approach has its strengths and weaknesses. One approach is the attempt to account rigorously for the underlying causes of uncertainty and how they emerge within real systems. Amongst the sources of uncertainty are modelling choices related to the level of model detail and the selection of system boundaries. Both affect simulation outcomes and model accuracy. Fundamental system attributes such as sensitivity to initial conditions, immeasurable or non-deterministic states, system/environment coevolution, and emergence of new behaviours also heavily contribute to uncertainty. Accounting for these differing contributions within a single unified approach is challenging. Most research attempting to address some of these challenges makes use of multi-agent systems. (e.g. see [5,6,7,8])

Fortunately, the dynamics of many complex systems display strong convergent behaviour [1,10]. Although there are innumerable paths to the future, the diversity of plausible future conditions that influence our decisions is likely to be much smaller. Furthermore, the impressive "awareness" of subject matter experts allows that many of these variable conditions can be articulated to some degree. Hence, one alternative to a unified, high fidelity modelling approach is to use domain knowledge to guide the investigation of important uncertainty attributes. In this study, we use this approach and highlight key issues relating to uncertainty and potentially influencing how we evaluate fleet options.

In the context of fleet development, relevant sources of uncertainty are: the types and frequency of critical missions that must be accomplished [9]; the hardware, tools, or systems that would be best suited or feasible to achieve each of the missions; and the economic environment within which the different options for fleet development are assessed. We perceive these to account for some of the most important aspects of uncertainty related to the external environment, internal capabilities, and the goals driving organizational action, respectively. In this work, we study models in which uncertainty is present only in the external environment. Investigation of the other forms of uncertainty and their interactions will be addressed in future work.

## 3. STOCHASTIC FLEET ESTIMATION

To evaluate fleet performance in an air transportation problem, we need to estimate the expected work the fleet will have to accomplish over some given period of time. For this purpose, we use the Stochastic Fleet Estimation (SaFE) algorithm [11]. SaFE is a Monte-Carlo based approach that generates average yearly task requirements from a dataset with frequency, duration, and capacity requirements for tasks and platforms.

### 3.1 Tasks and Platforms

To estimate the size and composition of a military air transportation fleet, we need to decide what type of tasks the fleet will need to perform and what platforms it will need to accomplish them. For example, the Canadian Air Force carries out tasks from the highly intensive Disaster Assistance Response Team (DART) mission, to carrying dignitaries, as well as search and rescue operations.

A sample dataset of 100 tasks for ten generic platform types was generated randomly. Each task has requirements for one or two cargo types (e.g., pax, loads), which each have between one and three subfunctions (e.g., VIP passengers, soldiers). Some platforms are suitable for some subfunctions and unsuitable for others. Platform capacities and operating costs are defined for each of the ten platforms. Each platform has two different capacities, one for each type of cargo. Individual platform capacities and operating costs are shown in Table 1.

Table 1: Example Platform Capacities and Costs

| Platform | Type 1 Capacity | Type 2 Capacity | Cost |
|---|---|---|---|
| P1 | 22 | 5,072 | 0.9199 |
| P2 | 8 | 29,505 | 1.9282 |
| P3 | 9 | 95,467 | 5.6292 |
| P4 | 23 | 111,716 | 20.2813 |
| P5 | 58 | 25,812 | 2.8454 |
| P6 | 47 | 106,710 | 5.4026 |
| P7 | 100 | 23,827 | 4.7058 |
| P8 | 132 | 83,918 | 7.7959 |
| P9 | 200 | 21,165 | 3.4486 |
| P10 | 180 | 73,303 | 4.7107 |

Also, platform-specific durations are defined for each mission. The time it takes for one platform to accomplish its part of a mission primarily depends on the distance it has to travel. So each mission was randomly assigned one of four "duration-types", which are indicative of different mission distances. For each of the duration-types, platforms were assigned duration-bounds based on their capacities. Platform-specific durations were then drawn from a uniform distribution, within the duration-bounds on each platform. An example mission with subfunction requirements and platform-specific durations can be seen in Table 2. For example, platform P5 is only capable of performing subfunction 3 of cargo type 1, and takes 12.53 days to complete up to 58 units of this subfunction requirement.

### 3.2 Platform to Mission Assignment

For each task, there are multiple combinations of platforms that can be assigned to perform the task. The Platform to Mission Assignment (PTMA) algorithm [12] conducts an exhaustive search of all the possible combinations, and returns the valid platform assignments for each mission.

Although individual tasks are fairly constrained in the number of valid platform assignments, the space of possible optimal platform combinations is extremely large, with approximately $10^{101}$ possible solutions. Therefore it is computationally impossible to evaluate and analyse every point in the solution space.

### 3.3 SaFE Model

As indicated earlier, the SaFE model is a Monte-Carlo simulation of the various tasks that need to be accomplished. First, we assume that the tasks are performed over some user-predefined number of discrete time intervals $T$. Second, the frequency, $f_i$, and duration, $d_i$, of task $i$ are drawn from a triangular distribution. If the task frequency has a fractional component, then a random number, $r$, is drawn and compared with $f_i$ mod 1, to determine if the task will be carried out: if $r \leq f_i$ mod 1, then replace $f_i$ with $\lceil f_i \rceil$, else with $\lfloor f_i \rfloor$. Next, the task instance start time $t_i$ is randomly selected such that $0 \leq t_i \leq T$. The platform use, $u$, for the period $[t_i, t_i+d_i]$ is then updated: $u_{t \in [t_i, t_i+d_i]} = u_{t \in [t_i, t_i+d_i]} + 1$. This is repeated for all tasks in the given time interval and for all required iterations or years $Y$. Finally, the average platform use over all iterations is given by

$$u = \frac{1}{Y} \sum_{j=0}^{Y} u^j \tag{1}$$

Table 2: Example Mission Requirements

| Cargo Type | Subfunction | Requirement | Platform Duration (in days) for platform type | | | | | | | | | |
|---|---|---|---|---|---|---|---|---|---|---|---|---|
| | | | P1 | P2 | P3 | P4 | P5 | P6 | P7 | P8 | P9 | P10 |
| 1 | 1 | 401 | 0 | 0 | 0 | 0 | 0 | 0 | 0 | 9.45 | 0 | 0 |
| 1 | 2 | 470 | 0 | 0 | 0 | 0 | 0 | 0 | 0 | 9.45 | 0 | 0 |
| 1 | 3 | 170 | 0 | 9.88 | 0 | 0 | 12.53 | 0 | 0 | 9.45 | 0 | 0 |
| 2 | 1 | 20038 | 13.96 | 9.88 | 0 | 0 | 0 | 0 | 0 | 9.45 | 0 | 0 |
| 2 | 2 | 20152 | 13.96 | 9.88 | 0 | 0 | 0 | 0 | 0 | 9.45 | 0 | 0 |
| 2 | 3 | 518347 | 0 | 0 | 0 | 0 | 0 | 0 | 0 | 9.45 | 0 | 0 |

and its corresponding standard deviation is designated by $\sigma_u$. Here, $u^j$ denotes the arithmetic time average of platform use for the interval $[0,T]$ at iteration $j$.

### 3.4 Scenarios

For the purposes of this paper, a scenario is defined as one year of simulated air transportation fleet activity. Depending on how task frequencies are generated (i.e., from the same or different distributions), the differences between the simulation years will be more or less accentuated. Key differences only come into play if the existence of major tasks or missions varies across the different scenarios. There is a high likelihood for this to happen since the execution of large tasks is usually not required very often (in the real world problem).

## 4. MULTI-OBJECTIVE OPTIMISATION

In order to generate optimal fleets with respect to platform cost and mission duration, a multi-objective optimisation (MOO) approach is employed. We apply the Non-dominated Sorting Genetic Algorithm–II (NSGA-II) [13] with a few modifications to adapt it to the fleet-mix problem. NSGA-II is an elitist evolutionary algorithm which groups individual solutions into non-dominated fronts, and also contains a crowding-distance assignment operator to preserve diversity in the population of solutions. We refer to [13] for details on NSGA-II.

### 4.1 Crossover and Mutation

New candidate solutions (offspring) are generated by applying the crossover operator to individuals selected from the parent population and then subjecting the resulting offspring to the mutation operator.

The individuals in the solution space are matrices *P*, with the same number of rows as the number of tasks in the problem, and the same number of columns as the number of platforms. Each row contains a platform assignment option for a different task, which is simply a row vector indicating the number of each type of platform to be used to complete the task. Since the structure of the individuals is atypical, custom crossover and mutation operators were necessary for this problem.

To create a child solution, $Q_i$, with the crossover operator, first the parents are chosen using a binary-tournament selection, see [13] for details. The child $Q_i$ then undergoes a mutation, which is determined by the mutation rate. For each task and based on the mutation rate, it is randomly decided whether or not the platform assignment is mutated. If the assignment is to be mutated, a different option is randomly chosen from all the valid platform assignments and assigned to the child.

### 4.2 Objective Fitness Functions

One of the main design parameters of a genetic algorithm is the choice of optimised cost functions. In this problem, we optimise for both monetary cost and total mission duration time. The monetary cost function that we use in this problem is defined as follows [14]:

$$E_{monetary} = \sum_{v} \left[ \sum_{i} (d_i(v) \cdot p_i(v)) \right] \cdot c(v) \quad (2)$$

where *D* is the durations matrix with elements $d_i(v)$ representing the time required for one platform of type *v* to complete its share of all instances of task *i* in a given year. *P* is the solution matrix with elements $p_i(v)$ representing the number of platforms of type *v* that are needed to perform task *i*, $c(v)$ is the cost of using platform *v*. When combined with each task's platform-specific durations, the rows of *P* encode the task-required platforms, forming the desired fleet.

The total mission duration cost function that we use in this problem is defined as follows [14]:

$$E_{time} = \sum_{v} \sum_{i} (d_i(v) \cdot p_i(v)) \quad (3)$$

.In this case, the fleet configuration of platforms $\{v\}$, $\sum_{i} (d_i(v) \cdot p_i(v))$ represents the required usage of vehicles of type *v* in platform-days/day. The term "platform-days/day" refers to the amount of time it would take for a single platform of type *v* to complete an average day's required work. Minimising this function results in the totality of tasks taking on average less time to complete.

# 5. ROBUSTNESS VS. ADAPTABILITY

The multi-objective optimisation algorithm generates multiple optimal fleets that can accomplish a given scenario. Each scenario leads to a set of potentially different optimal fleets with respect to fleet configuration, optimal platform cost and mission duration. However, many scenarios can be accomplished by identical fleets. Fleets that can effectively accomplish a set of mission scenarios are said to be robust to that set of scenarios. To evaluate robustness we examine all fleets generated by the MOOs for all scenarios and determine the supersets of the fleets generated by any given scenario. If a fleet or a subset of this fleet is not present in the fleet solutions of another scenario, then it is assumed that the fleet cannot accomplish the given scenario.

In order to determine the adaptability of a fleet, we calculate by how many platforms this (failing) fleet needs to be augmented in order to accomplish a scenario it cannot complete in its current configuration. We then group the scenarios according to the additional funding required to make the fleet fully effective. We want to find out how well a fleet is strategically positioned, i.e. how well can it adapt to potential future scenarios. An adaptable fleet will only require a small infusion of money in order to address effectively a large number of distinct scenarios. It is unlikely that any cost-constrained fleet will be able to address every scenario. Scenarios that a fleet cannot address (adapted or otherwise) pose a unique risk to that particular fleet. Hence, each particular fleet configuration will have a unique profile of scenarios for which it is robust, adaptable, or risky. How a scenario is classified for a particular fleet depends on what adaptations are feasible. We can let a decision maker decide by how much the fleet can adapt based on cost increase thresholds. The results of our analysis and the threshold data would then enable us to decide how well a fleet can be *adapted* to possible future scenarios.

The algorithm for determining which scenarios a particular fleet can adapt to is given as follows:

1. Take $Y$ scenarios (from SaFE)
2. For each scenario:
   a. Run NSGA-II for $N$ individuals and $G$ generations
   b. Map the solutions (fleet assignments) for each scenario into fleets
   c. Remove any duplicate fleets
3. For each scenario: ($i = 1, ..., Y$)
   a. For each fleet:
      i. For each scenario: ($k = 1, ..., Y$)
         1. Compare the fleet to every fleet in the $k$-th scenario
         2. If the fleet contains one of the fleets in the $k$-th scenario, then the fleet can accomplish the $k$-th scenario
      ii. Assign the fleet a score based on the percentage of scenarios it can accomplish

# 6. RESULTS

Populations of $N=500$ individuals were optimised using NSGA-II on $Y=100$ scenarios which were generated with SaFE. Each genetic algorithm was run for $G=100$ generations, with a mutation rate of *0.25*, and after each run, the individuals were mapped to fleets. After removal of duplicate fleets within each scenario, there were *45,495* total fleets.

Two different experiments were run. Experiment 1 used 100 low-variability scenarios, and Experiment 2 used 100 scenarios with high-variability.

## 6.1 NSGA-II Results

Figure 1 shows sample results from one NSGA-II optimisation run. Dark blue points indicate solutions which are part of the non-dominated front, while dark red points are part of the last front. The points on the first front are joined with straight lines to aid the eye only. The fleet costs range from 200 to 290.

## 6.2 Fleet Capabilities: Experiment 1

Each fleet in each scenario was compared with the fleets in every other scenario to determine each fleet's capabilities. For a fleet $X$ of scenario $A$, if one of the fleets optimised for scenario $B$ is a subfleet of $X$, then $X$ is capable of performing $B$. Each fleet was assigned a fleet-capability score based on the percentage of scenarios which it could accomplish, for cost increases of up to 1%, 2%, 3%, 4%, and 5% of each fleet's original cost. The minimum cost required to increase each fleet such that it contained at least one other fleet in every other scenario were calculated, and fleet-capability scores were assigned to these increased fleets.

Figure 2 shows histograms of the fleet-capability scores for each of the above increases in cost. The fleets' capabilities seem to increase quite rapidly with relatively small increases in cost. With no cost increase, about 5,000 fleets could accomplish over 90% of the scenarios. With a cost increase of only 1%, over 90% of the scenarios could be adapted to by more than 10,000 fleets. With a cost increase of only 5%, over 90% of the scenarios could be adapted to by virtually all the fleets.

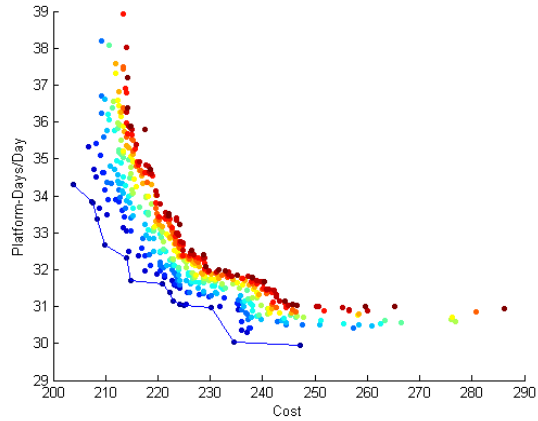

**Figure 1:** Objective scores of 500 individuals from one NSGA-II run

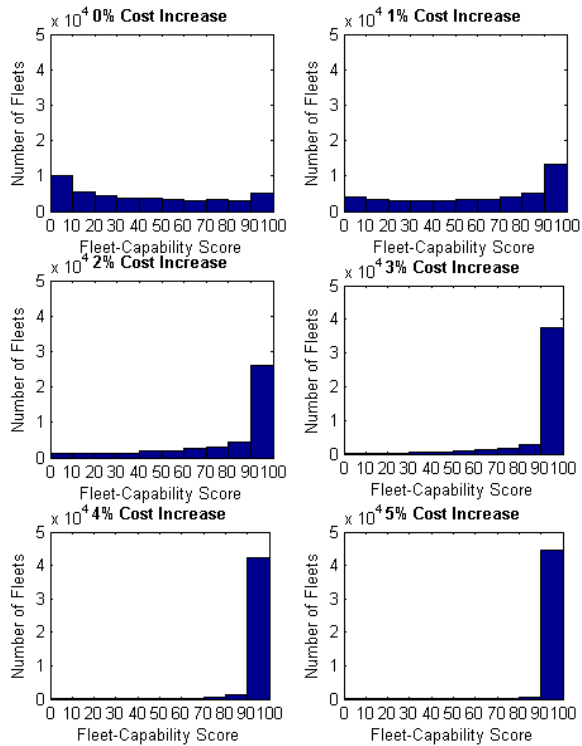

**Figure 2:** Fleet-capability score histograms for various cost increases on a set of fleets in Experiment 1

The required cost increase is so small because of the particular characteristics of the scenarios used. Both mission frequency and duration data were drawn from the same distributions, and since each scenario contains 100 missions, it is not very likely that the total platform demand over the scenarios has a high variability. Lastly, since a random population of fleets was optimised for each scenario, and the scenarios are similar, it is not surprising that there are similar fleets for each scenario.

### 6.3 Fleet Capabilities: Experiment 2

Experiment 2 used all of the same parameters as were used in Experiment 1 to create optimized sets of fleets on a group of scenarios. However, the scenarios themselves were altered before running the second experiment. For each scenario that was used in Experiment 1, a random number, *r, (1 ≤ r ≤ 2)*, was drawn from a uniform distribution. The mission frequencies for the scenario were then multiplied by *r*. Scenarios with a high multiplicative factor *r* will demand more from a fleet than scenarios with a low value for *r*. This multiplication increases the variance in demand among the scenarios. Fleet-capability scores were then calculated as in Experiment 1, except using different cost-increase levels, namely, 5%, 15%, 25%, 35%, and 50%. Figure 3 displays histograms of the fleet-capability scores for these cost-increase levels.

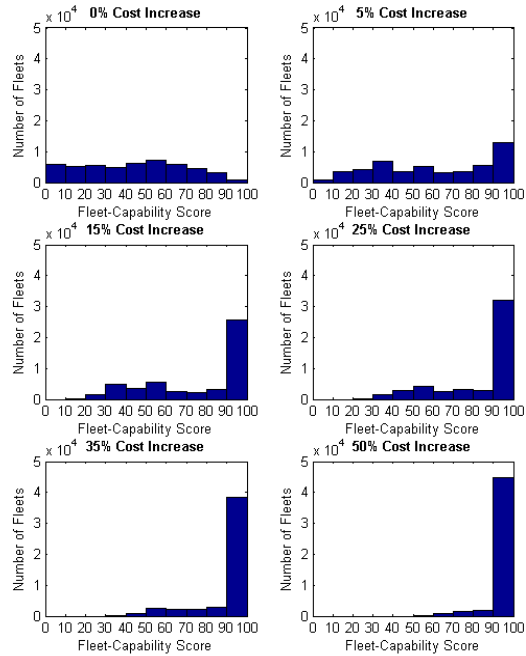

**Figure 3:** Fleet-capability score histograms for various cost increases on all fleets from Experiment 2

It can immediately be seen that the fleets from the second experiment are much less adaptable than those from the first experiment. Allowing a cost-increase level of 5% does not have nearly the same effect on the fleet's capabilities in Experiment 2 as in Experiment 1. This is because the demand of the scenarios in Experiment 2 varies much more than in Experiment 1. If a fleet $X$ was optimized on a scenario $A$ for which $r_A = 1$, it makes sense that it would take much more than a 5% increase in cost to allow $X$ to adapt to a scenario $B$ for which $r_B = 2$. When estimating the adaptability of any given fleet, it is important when testing the fleet's performance over various scenarios to use a realistic representation of all possible future scenarios.

### 6.4 Options for Fleet Growth

Each fleet has a variety of options for growth within a cost level. For example, if a fleet has a cost of 200, and if it can be increased by up to 1%, the fleet may be increased by either adding one of platform 1, two of platform 1, or one of platform 2 (see Table 1 for platform costs). An example fleet was chosen, and for each scenario that it could not already accomplish, the costs required to adapt the fleet were calculated. The costs of the different platform additions were organized according to the percentage of the original fleet's cost.

Figure 3 shows the number of options for increasing an example fleet's cost by up to zero, one, two and three percent in Experiment 1. Obviously, with a 0% increase in cost, nothing can be added to the fleet, so it must remain the same. For a cost increase of 1%, the fleet's capability can be increased by adding platforms in three different ways. These three new "one-percent increase" fleets have four, six, and thirteen options for growth which will add capability. Only four of the 23 total fleets with a 2% cost increase may be increased to higher-capability fleets within the 3% cost level.

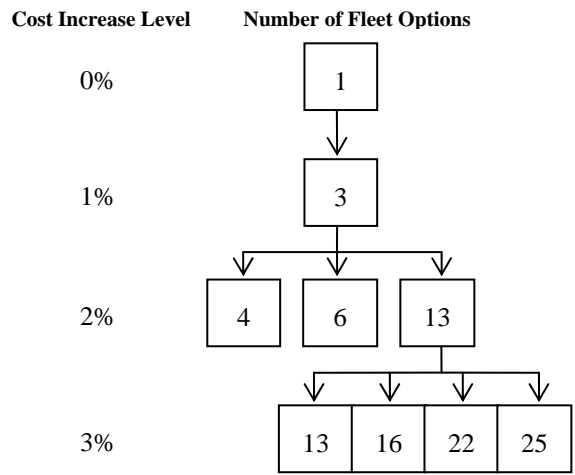

**Figure 3:** Number of options which an example fleet has for growth at each cost level

Increasing a fleet along different paths in Figure 3 will yield different higher-capability fleets which may be suitable for different scenarios. Adding certain platforms will add capability for a given set of scenarios, but not necessarily for other sets. It is important to understand how the original fleet may be augmented given possible budget increases, in order to respond to specific scenarios.

## 7. CONCLUSIONS

In this paper, we have shown how we could determine the degree of robustness and adaptability of a set of fleets. This information greatly helps in determining the strategic positioning of a particular fleet by examining its relationships with other potential fleets.

In the future we propose to analyse how an actual fleet can be augmented using this framework as well as to apply weights to each objective in order to capture the economic environment and appetite for risk of a decision maker. These important elements of risk will dramatically influence the perspective from which the different options for fleet development will be assessed.